\documentclass[10pt,twocolumn,letterpaper]{article}

\usepackage{cvpr}              % To produce the CAMERA-READY version
\definecolor{cvprblue}{rgb}{0.21,0.49,0.74}
\usepackage[pagebackref,breaklinks,colorlinks,allcolors=cvprblue]{hyperref}
\usepackage{cuted} 
\usepackage{amsmath}
\usepackage{makecell} 
\usepackage{siunitx}

\newcommand{\cameraready}[1]{{\color{black} #1}}

\def\paperID{34} % *** Enter the Paper ID here
\def\confName{CVPR}
\def\confYear{2025}

\definecolor{somegray}{rgb}{0.5, 0.5, 0.5}
\newcommand{\darkgrayed}[1]{\textcolor{somegray}{#1}}
\makeatletter
\newcommand*\titleheader[1]{\gdef\@titleheader{#1}}
\AtBeginDocument{%
  \let\st@red@title\@title
  \def\@title{%
    \vskip-3em
    \bgroup\normalfont\large\centering\@titleheader\par\egroup
    \vskip1.5em\st@red@title}
}
\makeatother

\titleheader{\darkgrayed{This paper has been accepted for publication at the\\
IEEE Computer Vision and Pattern Recognition Workshop (CVPRW), Nashville, 2025.
\copyright IEEE}}

\title{Reading in the Dark with Foveated Event Vision}

\author{Carl Brander \hspace{4mm} Giovanni Cioffi \hspace{4mm} Nico Messikommer \hspace{4mm} Davide Scaramuzza\\
Robotics and Perception Group, University of Zurich, Switzerland
}

\begin{document}
\maketitle
\begin{abstract}
% % Abstract from 07.03.25
% Current smart glasses with RGB-based cameras face challenges in low-light and high-motion environments. 
% % 
% They also pose excessive bandwidth and power needs, which reduce battery life. 
% % 
% We propose a novel event-based Optical Character Recognition (OCR) approach for smart glasses in low-light and high-motion environments. 
% % 
% Utilizing an egocentrically mounted event camera combined with the Meta Aria glasses' eye-tracking capability to foveate the event stream, we reduce bandwidth to a minimum. 
% % 
% We perform deep binary reconstruction trained on custom synthetic data and use multi-modal LLMs for OCR which improves the performance compared to traditional OCR solutions.
% % 
% % Using hierarchical image-stitching based on eye-gaze saccades, the images are sent to a cloud-based LLM to generate accurate OCR results.
% % 
% Our results demonstrate the ability to read text in dark environments down to moonlight brightness while using up to 2'400 times less bandwidth than a wearable RGB camera. 
% % 
% %Using a multi-modal LLM, we improve the performance of our approach compared to traditional OCR solutions.
% %Furthermore, we show the increased performance of multi-modal LLMs on coherent text recognition over traditional OCR solutions.
% % 
% %This event camera based OCR approach for smart glasses demonstrates the advantages of such systems in terms of reduced power consumption and bandwidth, especially in high-motion and low-light conditions.
% 
% 
% 
Current smart glasses equipped with RGB cameras struggle to perceive the environment in low-light and high-speed motion scenarios due to motion blur and the limited dynamic range of frame cameras. 
Additionally, capturing dense images with a frame camera requires large bandwidth and power consumption, consequently draining the battery faster.
These challenges are especially relevant for developing algorithms that can read text from images.
In this work, we propose a novel event-based Optical Character Recognition (OCR) approach for smart glasses. 
By using the eye gaze of the user, we foveate the event stream to significantly reduce bandwidth by around 98\% while exploiting the benefits of event cameras in high-dynamic and fast scenes.
Our proposed method performs deep binary reconstruction trained on synthetic data and leverages multi-modal LLMs for OCR, outperforming traditional OCR solutions.
% 
% Using hierarchical image-stitching based on eye-gaze saccades, the images are sent to a cloud-based LLM to generate accurate OCR results.
% 
Our results demonstrate the ability to read text in low light environments where RGB cameras struggle while using up to 2'400 times less bandwidth than a wearable RGB camera. 
\end{abstract}    
\section{Introduction}
\label{sec:intro}
The increase in popularity of wearable smart glasses featuring egocentric cameras has been fueled by the reduction in the size, weight, and power consumption of the required electronics~\cite{lin2025augmented, goesele25imaging}.
% 
% Commercial and research-focused smart glasses, as well as AR/XR devices, are some of the catalysts for egocentric computer vision.~\cite{Waisberg_Ong_Masalkhi_Zaman_Sarker_Lee_Tavakkoli_2023} 
% 
On the forefront of this increase in egocentric computer vision are commercial and research-focused smart glasses, as well as AR/XR devices ~\cite{Waisberg_Ong_Masalkhi_Zaman_Sarker_Lee_Tavakkoli_2023}.
% 
% Now that development hardware has become more widely available, various smart glass use cases are following suit in the research community.
% 
Since smart glass devices have become more widely available, various applications are emerging in the research community.
Areas such as egocentric Optical Character Recognition (OCR) to aid visually impaired people, as well as pipelines utilizing eye-gaze tracking to increase the efficiency of on-device algorithms, have been explored~\cite{mucha2024text2tasteversatileegocentricvision} ~\cite{wang2023gazesamsegment}.
% 

% 
% Although remarkable results are achieved on wearable devices, these approaches face major challenges due to the limitation of RGB cameras. 
% 
Recent research has shown remarkable results in these areas~\cite{Kim2021-io}.
However, challenges such as high battery consumption and low accuracy in low-light conditions hinder the deployment of these algorithms in consumer products. These issues are mainly due to the use of RGB cameras.

\begin{figure}[t]
    \centering
    %\vspace{0.92cm}
    \includegraphics[trim={0cm 0cm 0cm 0cm},width=\linewidth]{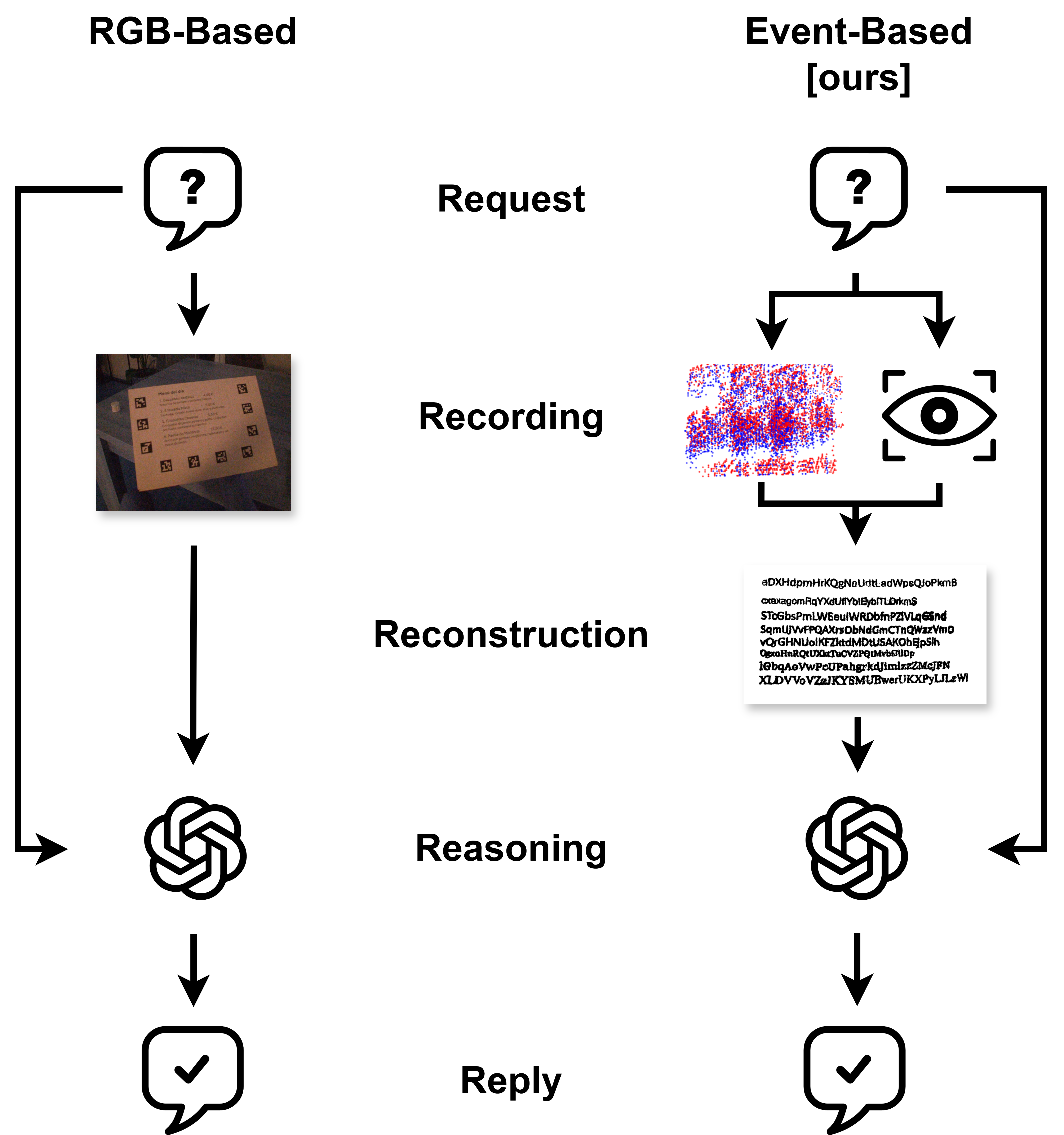}
    \caption{Example Process flow of a user query about a certain text in the scene. The query is answered by a cloud-based LLM system. On the left, is an existing RGB-based approach transmitting the full RGB image from the smart glasses to the cloud, and on the right, is our event-based pipeline performing reconstruction on the event stream and eye-gaze, then sending the reconstructed text to the cloud for OCR.}
    \label{fig:rgb_vs_event_overview}
\end{figure}

% RGB cameras are notoriously power-hungry for battery-powered wearable devices that struggle to power them continuously for more than a few hours at a time.~\cite{naderiparizi2018camerapowerhungry} While profiting largely from existing computer vision algorithms, their bandwidth scales proportionally with their framerate incurring large amounts of data to be processed.~\cite{Gehrig_Scaramuzza_2024} This not only uses more energy but also processing power which are both strongly limited on wearable devices.
% 
One main shortcoming of RGB cameras is their proneness to motion blur, low contrast and high image noise in low-light scenes~\cite{chakravarthi2024recenteventcamerainnovations}.
% 
% Combined with the fast rotational motion of the human head, the resulting blurry frames represent a difficult challenge for OCR.
% 
\begin{figure}[t]
    \centering
    %\vspace{0.92cm}
    \includegraphics[trim={0cm 0cm 0cm 0cm},width=\linewidth]{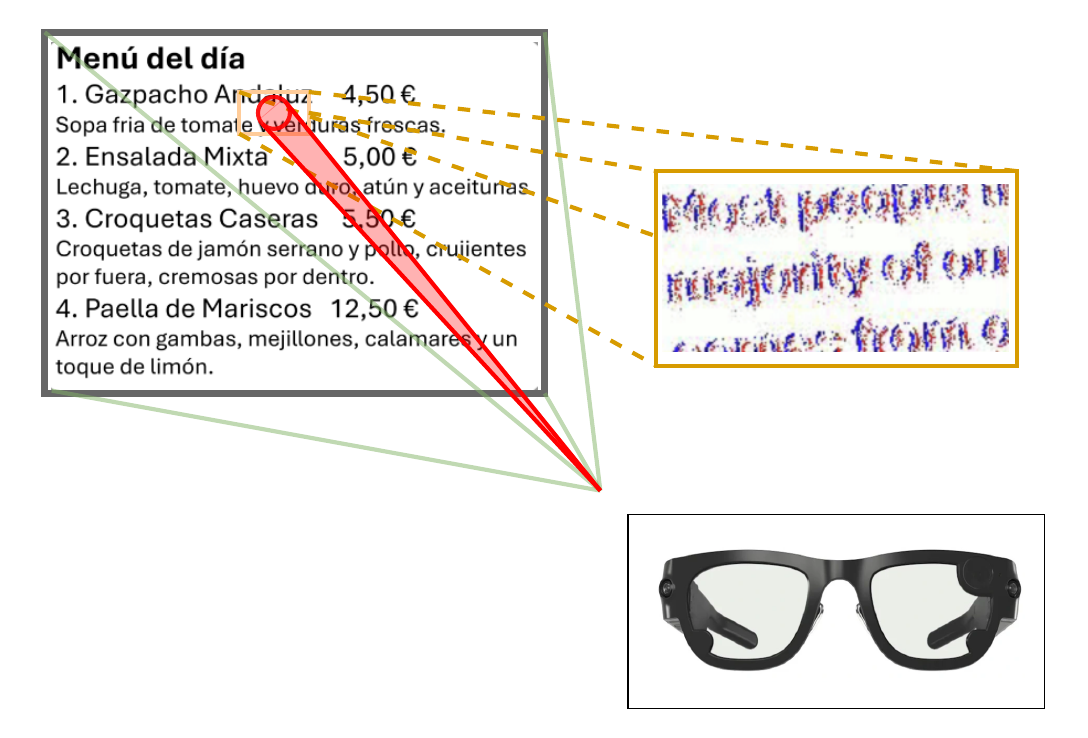}
    \caption{We use an event camera integrated on the Meta Aria glasses to perform object character recognition (OCR). Our event-based algorithm uses only the foveated events to perform binary image reconstruction and, then, OCR via large language models.}
    \label{fig:fig1}
\end{figure}
Another factor contributing to motion blur is the abrupt movement of the human head.
The resulting blurry frames represent a difficult challenge for OCR.
% 
% In recent years, key advances leveraging event cameras paradigm with higher resolutions and advanced computer vision models have explored new use cases~\cite{chakravarthi2024recenteventcamerainnovations}.
%
In recent years, key advances leveraging event cameras paradigm with higher resolutions and advanced computer vision models have explored new use cases such as egocentric motion capture or always-on human machine interfaces~\cite{millerdurai2024eventego3d3dhumanmotion}~\cite{bhattacharyya2024heliosextremelylowpower}~\cite{chakravarthi2024recenteventcamerainnovations}. 
% 
% These works have opened up the possibility of investigating event cameras as alternative solutions for various egocentric vision tasks such as OCR. 
%
% Event cameras possess great power efficiency while mitigating motion blur and drastically enhancing the dynamic range.~\cite{Gehrig_Scaramuzza_2024} 
% 
Event cameras have great potential to increase power efficiency while mitigating motion blur and drastically improving the dynamic range~\cite{Gehrig_Scaramuzza_2024} for egocentric vision tasks.
This makes them excellent extensions or even substitutions to the existing RGB cameras for egocentric vision tasks such as OCR. 
In this work, we investigate the benefits of egocentric event cameras to perform power-efficient and robust OCR for smart glasses. 
Our method is based on a multi-step process as depicted in Figure \ref{fig:rgb_vs_event_overview} that uses eye-tracking and a binary image reconstruction neural network. 
Our approach achieves up to 2'400 times reduction in bandwidth of the data to be transferred from the smart glasses to a cloud-based OCR engine or LLM when compared to a traditional RGB approach. 
% 
% Furthermore, it also extends the usage envelope of the smart glasses in low-light environments down to around moonlight brightness.
%
Furthermore, it enables OCR on smart glasses even in low-light conditions where RGB cameras struggle.
\section{Related Work}
\label{sec:related_work}
Wearable smart glasses have seen a rapid increase in popularity in the last few years.
Especially through the proliferation of customer-ready smart glasses through companies such as Meta, these devices have gained the interest of the broad population. Their market share is expected to grow rapidly in size within the next years, thanks to the increasing number of applications catered to by the smart glasses~\cite{Yutong_Hang_Chen_Ng_2021}.
This rise of smart wearable egocentric vision systems requires more and better egocentric, intelligent algorithms and machine learning models which have seen major improvements over the last years already~\cite{K._Rege_Zoltay-Paprika_Kumar_Ganie_2024}.
Chen and Duan show the use of egocentrically mounted RGB cameras to recognize MIDI music nodes~\cite{7477714}
While Wang \etal enables the segmentation of objects including the wearables eye-gaze to direct the model into a specific region of interest~\cite{wang2023gazesamsegment}.
Specific applications for egocentric OCR include Shenoy's \etal research titled \textit{LUMOS} enabling wearable vision system OCR with a cloud-connected Large Language Model to reason on the text seen in the image~\cite{shenoy2024lumosempoweringmultimodal}.
Similarly, Mucha \etal demonstrate the ability to read menu cards using the Meta Aria glasses' RGB camera to increase the independence of visually impaired people in their daily lives~\cite{mucha2024text2tasteversatileegocentricvision}.\\
Egocentric event cameras designated to be used on smart glasses have not seen this level of progress. Yet, there is increased interest in using event cameras on smart glasses for eye tracking as Feng \etal show in their research~\cite{feng2022realtimegazetrackingeventdriven}.
Or using event-based cameras as Human-Machine-Interfaces (HMI) for all-day online gesture recognition as Bhattacharyya \etal show~\cite{bhattacharyya2024heliosextremelylowpower}.
Using event cameras as substitutes to the prominent RGB cameras for scene understanding on smart glasses is also becoming more feasible through the increase in available datasets such as $E^2(GO)MOTION$ from Plizzari \etal presenting a large event-based action recognition dataset based on egocentric event camera data~\cite{plizzari2022e2gomotionmotionaugmentedevent}.
And Millerdurai \etal proposing methods to capture human motion with egocentrically mounted event cameras in \textit{EventEgo3D}~\cite{millerdurai2024eventego3d3dhumanmotion}.
Finally, Wang \etal recently presented a fully event-based OCR pipeline using transformers including a novel event-based scene text recognition dataset called \textit{EventSTR}~\cite{wang2025eventstrbenchmarkdatasetbaselines}.
In comparison, while Wang \etal also substitutes RGB cameras for event-based cameras and performs OCR using LLMs, our approach additionally implements both foveation and binary reconstruction to address the high bandwidth required by RGB cameras while still being able to use off-the-shelf image-based OCR software. This can be seen in Table \ref{tab:related_works}. It shows a comparison of covered topics in existing literature compared to our findings.
\renewcommand{\arraystretch}{1.5}
\begin{table}[h]
\caption{Covered topics of selected related works.}
\label{tab:related_works}
\resizebox{\columnwidth}{!}{%
\begin{tabular}{l|c|c|c|c|c|c}
\makecell{Covered Topics / \\ Selected Related Works} & 
\makecell{~\cite{7477714}} &
\makecell{~\cite{wang2023gazesamsegment}} & 
\makecell{~\cite{shenoy2024lumosempoweringmultimodal} \\ ~\cite{mucha2024text2tasteversatileegocentricvision}} & 
\makecell{~\cite{feng2022realtimegazetrackingeventdriven} \\ ~\cite{bhattacharyya2024heliosextremelylowpower}\\ ~\cite{plizzari2022e2gomotionmotionaugmentedevent} \\ ~\cite{millerdurai2024eventego3d3dhumanmotion}} & 
\makecell{~\cite{wang2025eventstrbenchmarkdatasetbaselines}} &
\textbf{ours}\\ \hline \hline
Egocentric OCR with RGB Cameras   &
\checkmark       &
                 &               
\checkmark       &         
                 &
                 &
\checkmark\\ \hline

Egocentric Event-Based Cameras          &
                                        &                                             
                                        &                          
                                        &  
\checkmark                              &
                                        &
\checkmark \\ \hline

Event-Based OCR                         &                          
                                        &                                              
                                        &                          
                                        &
                                        &
\checkmark                              &
\checkmark                \\ \hline

Foveation using Eye-Gaze Tracking       &                          
                                        &     
\checkmark                              &                      
                                        &  
                                        &
                                        &
\checkmark   \\ \hline         

\end{tabular}%
}
\end{table}
\section{Methods}
\label{sec:methods}
\subsection{Overview}
\label{sec:overview}
\begin{figure*}[th]
    \centering
    \includegraphics[trim={0.5cm 0cm 0.5cm 0.5cm},width=1\textwidth]{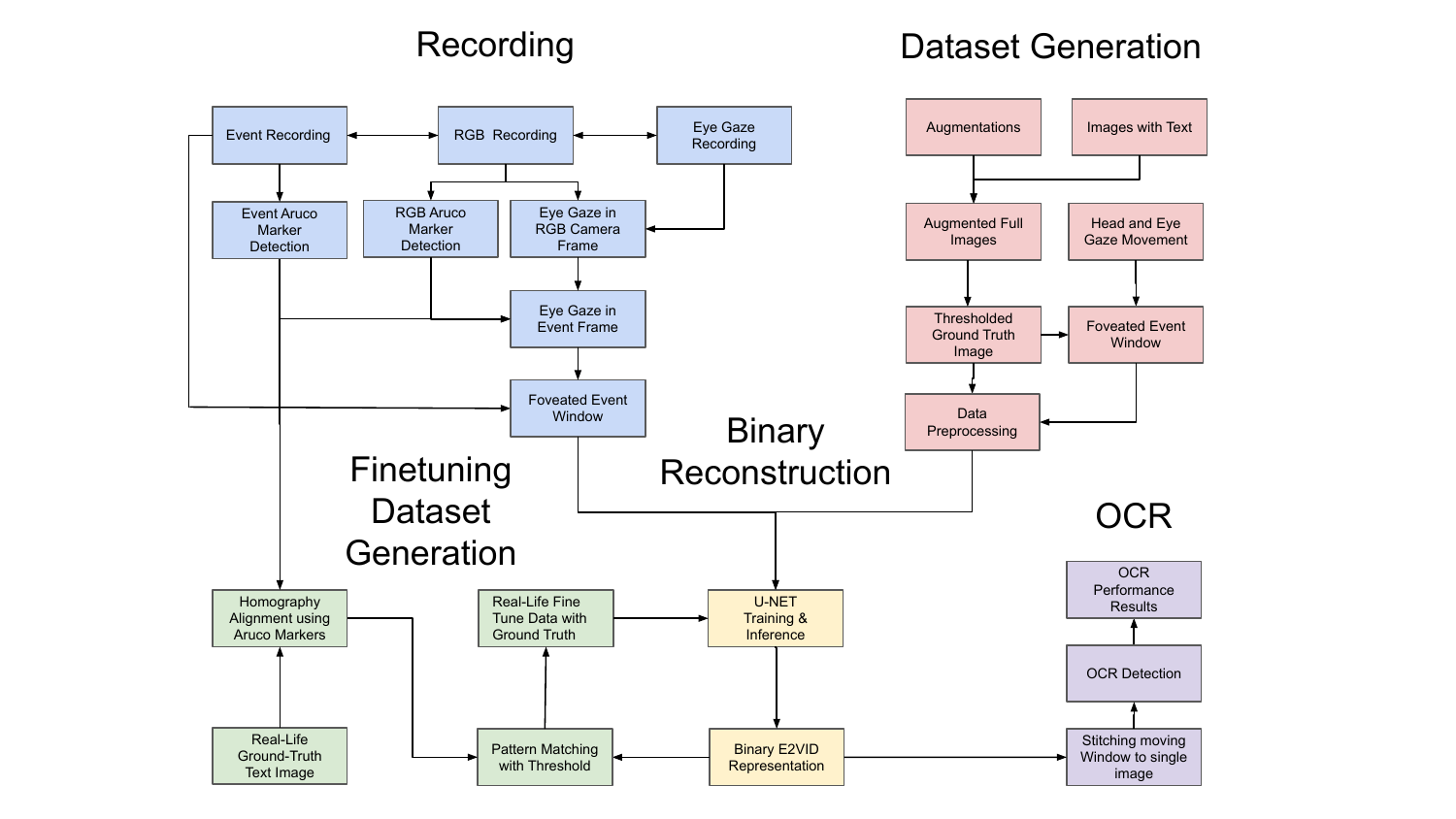}
    \caption{Event-Based pipeline structure split into the five main elements by color. All are computed offline after real-life data capture.}
    \label{fig:pipeline}
\end{figure*}
Figure \ref{fig:pipeline} visualizes the key steps required in data acquisition and processing. Section \ref{sec:data_acquisition} details the blue part of Figure \ref{fig:pipeline} such as recording and spatiotemporally aligning data streams in real life. Section \ref{sec:synthetic_data_acquisition} analyzes the synthetic data generation in the red part of Figure \ref{fig:pipeline} including augmentation and video-to-event transformation. Furthermore, Section \ref{sec:fine_tuning} lais out how real-life event-stream recordings are annotated with digital ground truth data in an automated fashion as seen in Figure \ref{fig:pipeline} in green. \\
Section \ref{sec:model_training} explains how the binary reconstruction model, used in the yellow part of Figure \ref{fig:pipeline}, was trained. Finally, Section \ref{sec:data_processing} details how the reconstructed time series of images are stitched together and OCR is performed as seen in the purple part of Figure \ref{fig:pipeline}.

\subsection{Data Acquisition}
\label{sec:data_acquisition}
Due to the unavailability of wearable devices with integrated event cameras and eye-tracking capabilities, it was required to mount the Metavision IMX636 event-based sensor onto the Meta Aria glasses. % as seen in Figure \ref{fig:hardware}.
%\begin{figure}[thb]
%    \includegraphics[trim={15cm 15cm 15cm 15cm},clip,width=0.5\textwidth]{images/Hardware.png}
%    \caption{Metavision IMX636 event based Sensor~\cite{PROPHESEE_2025} in custom housing including a 1/2" 4-12mm manual focus lens mounted on the Meta Aria smart glasses using a custom, rigidly fixed, 3D printed adapter bracket.}
%    \label{fig:hardware}
%\end{figure}
%
While the human fovea has an high 60 pixel/degree angular resolution, the Meta Aria glasses' RGB camera angular resolution in high-resolution mode producing 2880x2880 pixel images at a 110° FoV only manages $\approx$~26 pixel/degree~\cite{engel2023projectarianewtool}~\cite{deering1998limits}. To match the RGB camera's angular resolution, the IMX636 event-based sensor's FoV was tuned using the adjustable lens to be 50° in HFOV and 28° in VFOV resulting in an angular resolution of $\approx$~26 pixel/degree with it's HD resolution of 1280x720 pixel. This allowed for an unbiased comparison of OCR performance at various distances without one of the sensors having an advantage in angular resolution. 
Furthermore, the two data streams needed to be both spatially and temporally aligned within reasonable accuracies for eye-gaze foveation. This would not be required if commercial smart glasses with an integrated event camera were available, as this would solve both spatial and temporal alignment between eye-gaze and event stream without the following tedious method.\\
Temporal alignment within $\pm{~8ms}$ was achieved using a series of flashing Aruco markers on a computer screen at the start of each recording session~\cite{Garrido-Jurado_Muñoz-Salinas_Madrid-Cuevas_Marín-Jiménez_2014}. They were detected by both the Meta Aria glasses' RGB camera and the event camera recording simultaneously. The appearance of each Aruco Marker in each data stream was timed. The event stream could then be temporally aligned with the Meta Aria glasses' internal timestamps enabling synchronization between the eye-gaze tracking and the event stream.\\
As for the spatial alignment, multiple static Aruco markers were placed randomly around the text to be read.  This enabled the Meta Aria glasses' RGB camera and the event camera to detect all markers simultaneously, using the Event to Grayscale reconstruction model E2VID for the event-camera~\cite{Rebecq19cvpr}. Therefore, allowing the computation of a homography transformation assuming the text object to be a plane in 3D space, and the subsequent translation of the RGB and eye-gaze data into the coordinate frame of the event camera.\\
The inbuilt eye-gaze calibration routine of the Meta Aria glasses was used to increase the precision of eye-gaze tracking at the start of each recording session.
Once the spatial and temporal alignment, as well as eye-gaze calibration, was completed, each recording session featured reading a black-on-white printed text ranging from random letters to random words, normal text paragraphs, or famous pangrams in different environment conditions such as low-lighting or increased motion.

\subsection{Synthetic Data Acquisition}
\label{sec:synthetic_data_acquisition}
\begin{figure}[thb]
\includegraphics[width=\linewidth]{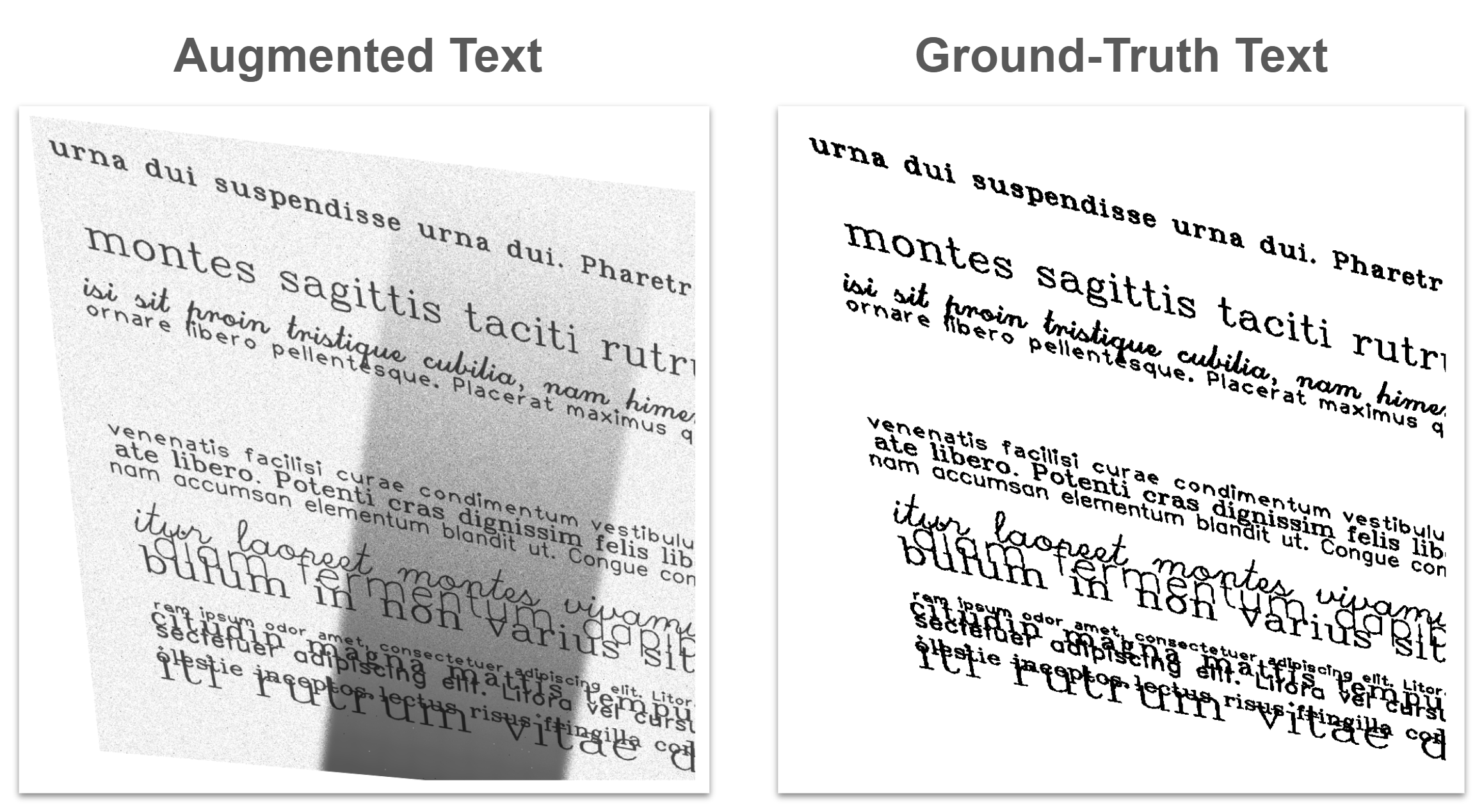}
\caption{Synthetically generated and augmented text example on the left next to the binary ground truth on the right before eye-gaze simulation and subsequent foveation.}
\label{fig:synthetic_text}
\end{figure}

Synthetic multi-modal reading data was acquired to pretrain the binary reconstruction network to perform a transformation of the event stream voxel input to a binary black-and-white image of the foveated text areas.
It was generated in large quantities through emulating eye, head, and hand movement in 6 Degrees of Freedom on a random text sample augmenting both movement, visual qualities, and 3D transformations in space as key factors for diversity in the dataset. An example of a warped and augmented text next to the binary ground truth can be seen in Figure \ref{fig:synthetic_text}. Starting from this augmented text passage, the eye-gaze movement was superposed to foveate the area into a 100x200 pixel-sized region of interest of the randomized eye-gaze movement. The generated simulation of eye-gaze reading a text paragraph was translated from grayscale frames into a realistic event stream using VID2E from Gehrig \etal~\cite{Gehrig_2020_CVPR}.
This method produced realistic data at high framerates of 4'000 frames per second thanks to its synthetic nature. Thus increasing the realism of the derived event-stream data without the need for interpolation.

\subsection{Finetune Dataset Generation}
\label{sec:fine_tuning}
\begin{figure}[htb]
    \centering
    \includegraphics[width=\linewidth]{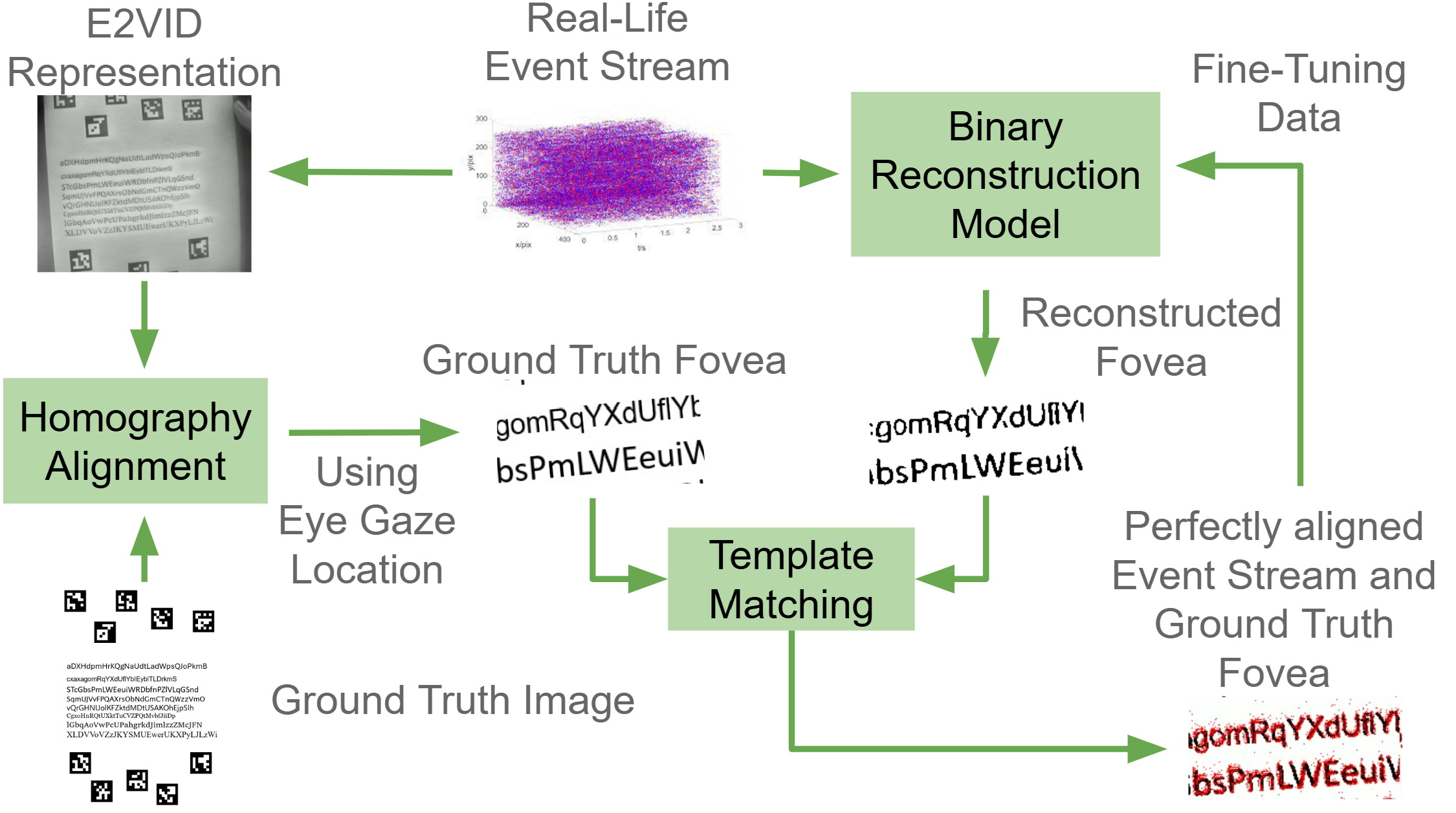}
    \caption{In-Detail layout of Fine-tuning Pipeline showcasing continued learning utilizing the neural networks output to create automatically binary annotated real-life event stream data for fine-tuning.}
    \label{fig:fine_tuning_pipeline}
\end{figure}

To increase the performance of the synthetically pretrained binary reconstruction model with the aim to transform the event stream voxels into binary images, real-life data was processed and spatiotemporally aligned with the available ground truth vector data of the real-life text paragraphs to fine-tune the model. This process is outlined in Figure \ref{fig:pipeline} in green color and described in detail in Figure \ref{fig:fine_tuning_pipeline}. To align the digital ground-truth data with the captured event stream, the E2VID transformation network was used again to detect the static Aruco markers along the text and align them with the markers on the ground-truth using homography estimation~\cite{Rebecq19cvpr}. To account for any temporal or spatial misalignments, pattern matching was used to minimize the misalignment that could occur due to the up-to $\pm$~8ms temporal misalignment between event stream and the Meta Aria glasses data-streams.  The already synthetically pre-trained binary segmentation neural network was leveraged to generate binary images of the real-life event stream which were aligned to the ground truth with above mentioned pattern-matching algorithm. This resulted in near-perfect annotation of real-life data available to be used for fine-tuning. Furthermore, it created a positive reinforcement loop enabling the model to increase its performance continuously using its outputs as seen in detail in Figure \ref{fig:fine_tuning_pipeline}.
\subsection{Model Training}
\label{sec:model_training}
\begin{figure}[htb]
    \centering
    \includegraphics[width=\linewidth]{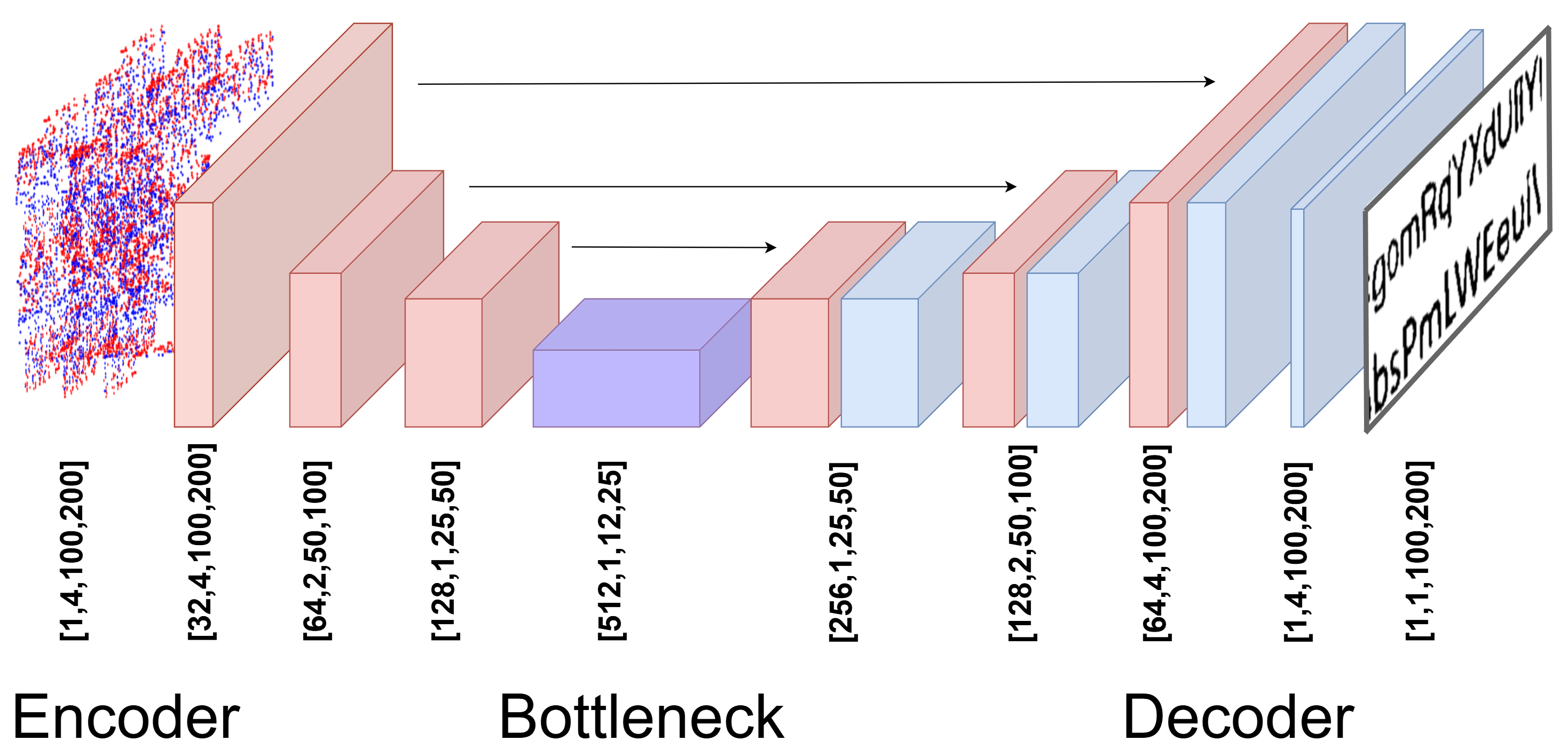}
    \caption{Voxel-adapted, U-Net based, binary reconstruction neural network architecture to transform a voxelized and discretized event stream into binary black-and-white images within their foveated 100x200 pixel spatial dimensions.}
    \label{fig:neural_network}
\end{figure}

A model transforming a voxelized event stream into binary black-and-white images destined to run on wearable devices is required to be space and power-efficient while not introducing too much latency. This was achieved using an efficient feed-forward architecture based on the U-Net structure first introduced by Ronneberger \etal~\cite{ronneberger2015unetconvolutionalnetworksbiomedical}. The structure can be seen in Figure \ref{fig:neural_network}.\\
This NN was first pretrained on $\approx$ 90'000 synthetic data points each consisting of a ground-truth binary foveated 200x100 pixel image and a 3D voxel-grid containing the past 1'600 Events temporally equally spaced into 4 voxel bins. After pretraining, real-life data fine-tuning was performed using the fine-tuning data gathered as described in Section \ref{sec:fine_tuning} using a learning rate reduced by $10^3$ compared to the initial learning rate.
A binary segmentation cross-entropy loss was used during training as well as a thresholding layer at inference to enforce hard predictions.

\subsection{Data Processing}
\label{sec:data_processing}
\begin{figure}[htb]
    \centering
    \includegraphics[trim={0.5cm 0.2cm 0.5cm 1cm},width=1\linewidth]{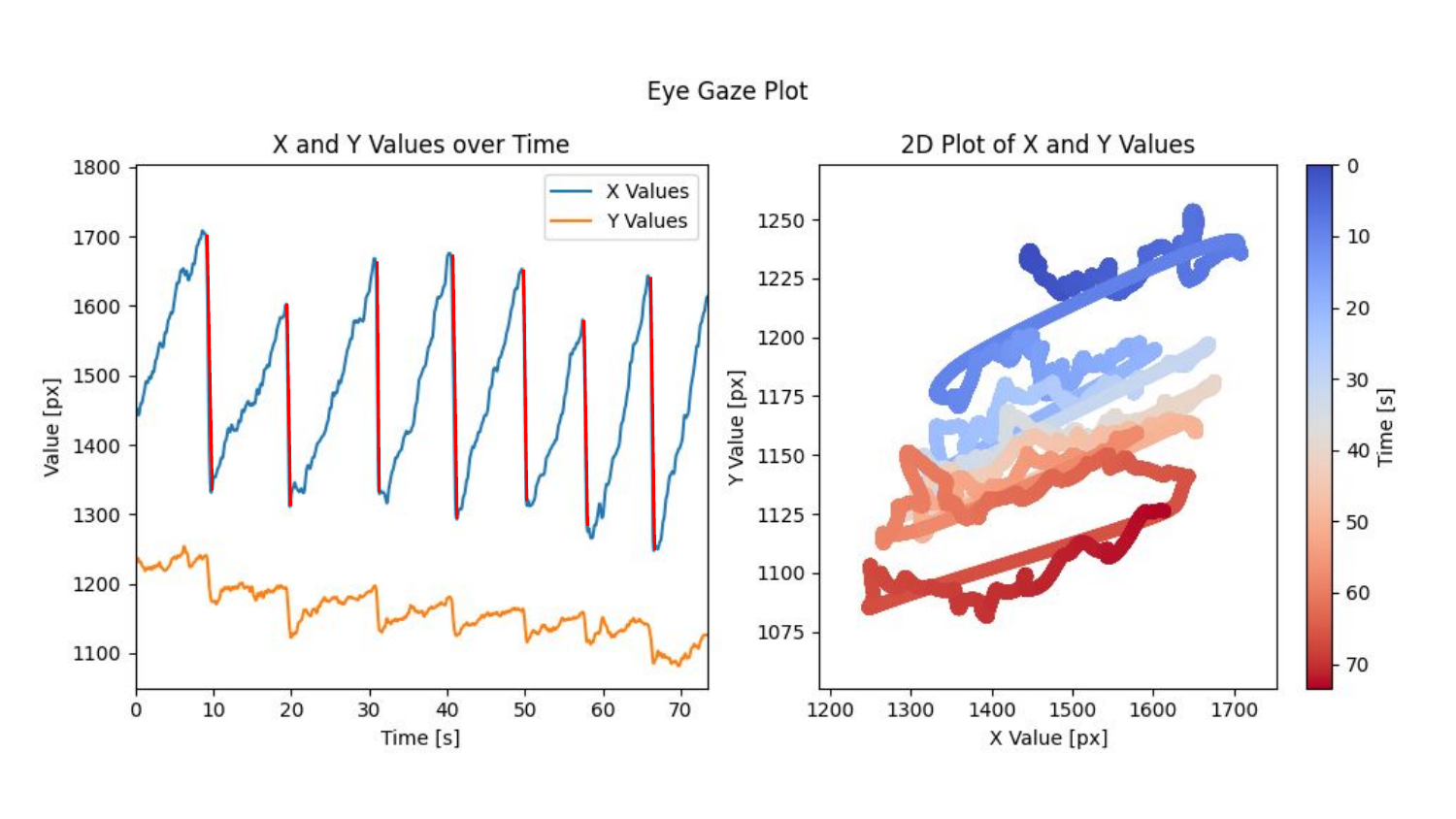}
    \caption{Left Diagram showing X and Y values of the user's eye-gaze over time when reading a paragraph of text. In red the eye-gaze saccades between different lines of text can be seen signified by their large reduction in X coordinate location.  \\
    The Right plot visualizes the eye-gaze location while reading on an X-Y space color-coded by time.}
    \label{fig:eye_gaze_plots}
\end{figure}

This Section summarizes the data flow through data acquisition (Blue) and binary reconstruction (Yellow) and explains the final part of the pipeline (Purple) which is hierarchical image stitching and OCR. Once temporally and spatially aligned eye-gaze and event stream data of the real-life recording is available, it is foveated to a region of interest based on the user's eye gaze to simulate the human vision fovea. This reduces the event stream's bandwidth by $\approx$ 98\% which in turn drastically reduces processing redundant data. This foveated event stream window of size 100x200 pixels is fed into the binary reconstruction neural network as presented in Section \ref{sec:model_training}. It outputs a time series of 100x200 pixel binary images resembling the text visible to the user's fovea while reading the text paragraph.
This sequence of binary images is fed into a hierarchical image stitching pipeline combining the individual frames based on their overlap and utilizing an alpha channel to average out temporal inconsistencies. An improvement mask, tracks pixel-wise areas of highest correlation between any past stitched frames to enforce high-quality additions and discards low correlation frames on a pixel-wise basis.
The update rule used is described in Equation \ref{eq:improvement_mask} and is used in Equation \ref{eq:stitched_image} to update the stitched image.
\begin{equation}
M_{imp}^{(x,y)} =
  \begin{cases} 
      M_{new}^{(x,y)}, & M_{new}^{(x,y)} > M_{saved}^{(x,y)} \\
      M_{saved}^{(x,y)}, & otherwise \\
   \end{cases}
  \label{eq:improvement_mask}
\end{equation}\\
\begin{equation}
I_{stitched}^{(x,y)} =
  \begin{cases} 
      F_{new}^{(x,y)}, & M_{new}^{(x,y)} > M_{saved}^{(x,y)} \\
      I_{stitched}^{(x,y)}, & otherwise \\
   \end{cases}
  \label{eq:stitched_image}
\end{equation}
Furthermore, it leverages hierarchical stitching based on the detection of eye-gaze saccades between lines of text read to reduce the search space of the stitching algorithm and increase efficiency as seen in Figure \ref{fig:eye_gaze_plots}.\\
This also combats misalignments due to possible 6 Degree of Freedom rotations of the text during reading by utilizing KLT Tracking \cite{lucas:hal-03697340} to stitch the individual text lines together essentialy de-warping them with their estimated transformation.\\
Once a stitched and thresholded binary reconstruction of the text is created, it resembles a downsized, binary image available to be used by traditional image-based OCR algorithms to digitize the text. This was done using either a dedicated OCR API from Google Cloud~\cite{Google} or an API leveraging OpenAI's LLM GPT-4o-2024-05-13~\cite{ChatGPT-4o}.
\section{Results}
\label{sec:results}

\subsection{Setup}
\label{sec:label}
The results were generated in different sessions using various text examples, Aruco Marker placements, and under different background illumination and movement settings. 
Before reading, the Meta aria glasses' eye-gaze calibration has been run to increase the glasses' eye-gaze tracking accuracy. The texts were printed in black with varying sizes, thicknesses, and font types on plain white paper which was supported not to bend during reading representing a flat plane. The room brightness was measured in Lux recorded from two separate devices of which the average was used to report brightness in the following figures.

\subsection{Egocentric OCR Results}
\label{sec:egocentric_ocr_results}
\begin{figure}[htb]
    \centering
    \includegraphics[trim={1cm 0cm 1cm 1cm}, width=\linewidth]{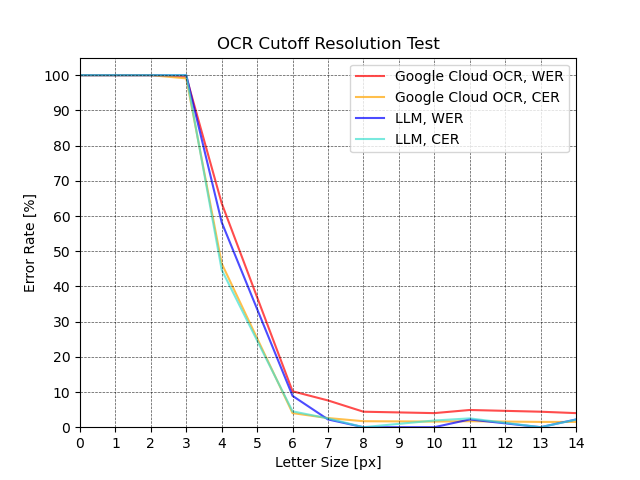}
    \caption{Word error rate (WER) and Character error rate (CER) over different small-case letter heights measured in pixels of an example non-augmented, non-warped text image.}
    \label{fig:ocr_cutoff_resolution}
    \vspace{-0.5cm}
\end{figure}
In Figure \ref{fig:ocr_cutoff_resolution}, GPT-4o-2024-05-13 and Google Cloud OCR's performance is measured by the WER (Word Error Rate) and CER (Character Error Rate) on a non-augmented, non-warped black-and-white text. At a threshold of 6-7 pixels in letter height of a small-case non-bold letter, both OCR approaches start steeply increasing their WER and CER. Therefore, letter sizes under a height of $\approx$~6 pixels lead to both models misinterpreting large amounts of the text. This 6-pixel threshold depends on the distance of the text to the camera, the camera's resolution, and its FoV (Field of View).\\
This result is directly correlated with the required angular resolution of both the event-based and the RGB camera required to perform error-free OCR when egocentrically mounted on smart glasses. 
Based on our experiments in Section \ref{sec:egocentric_ocr_results} we conclude the importance of angular resolution for OCR applications, agnostic of camera topology. With an average reading distance of $\approx$~50cm between camera and text and a common lower-case letter height of 2.25mm (for 12pt letters) our current wearable cameras such as the Meta Aria glasses with an angular resolution of $\approx$~26 pixel/degree manage to get resolution of $\approx$~6.7 pixel per letter, barely enough to keep the text legible to OCR algorithms as seen in Figure~\ref{fig:ocr_cutoff_resolution}. Whereas the human fovea with $\approx$~60 pixel/degree angular resolution at 20/20 vision and the same assumed cutoff resolution per letter, can read the same text at up to 1.2m distance~\cite{deering1998limits}. Therefore, we find that higher angular resolution image sensors for both event-based and RGB cameras are beneficial for wearable OCR applications in the future. Yet, current imaging sensors fitting the requirements of low-power consumption and small size are only marginally suitable for OCR on smart glasses right now.

\subsection{Performance Results}
\label{sec:performance_results}
\begin{figure}[h]
    \centering
    \includegraphics[trim={0cm 1cm 0cm 1cm},width=\linewidth]{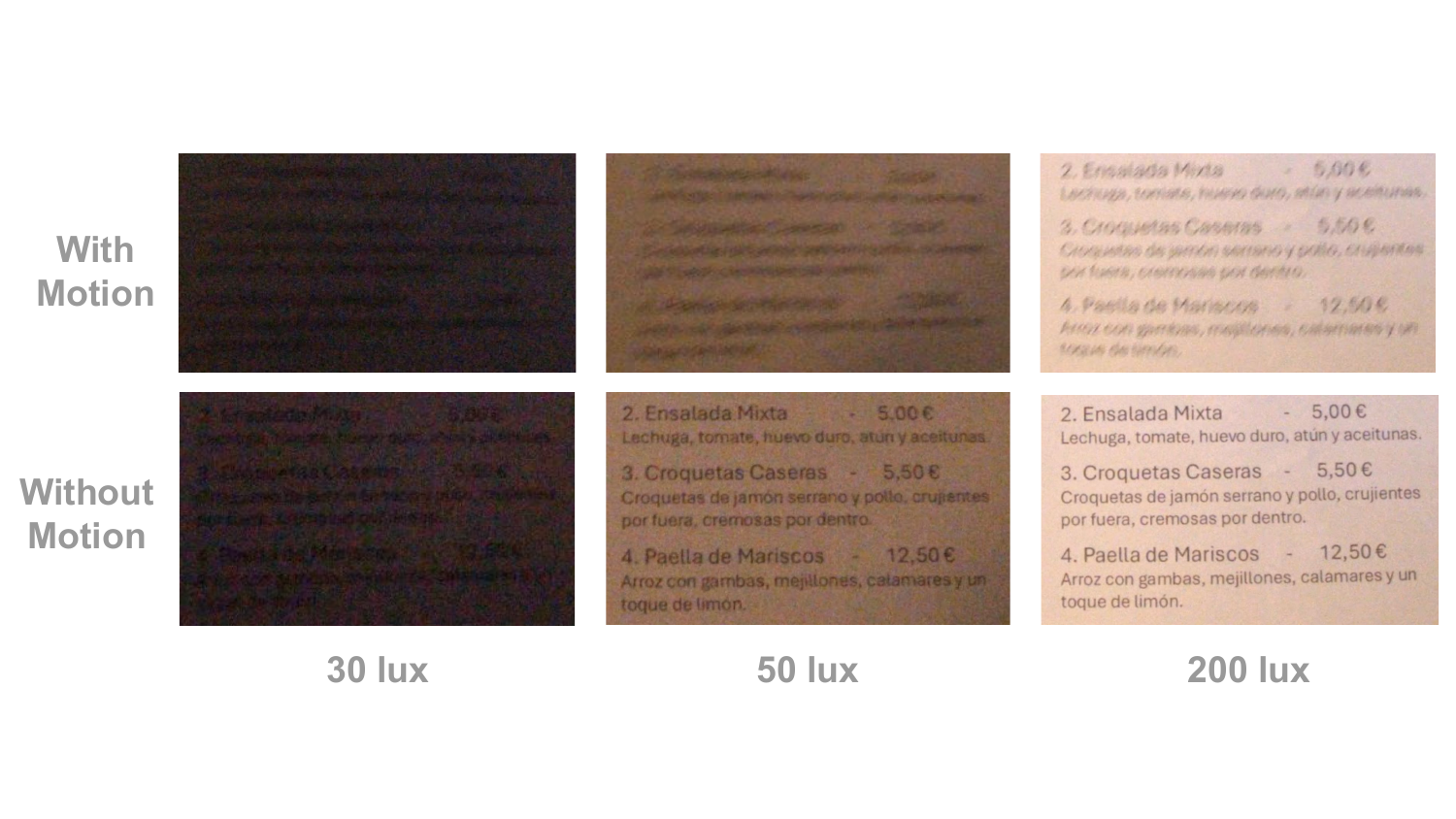}
    \caption{Meta Aria Glasses RGB Camera Image examples for OCR at different brightness levels in combination with motion.}
    \label{fig:rgb_in_low_light}
\end{figure}

The OCR performance was compared between the Meta Aria glasses' high-resolution RGB camera in a single high-resolution image snapshot and our continuous binary reconstructed and foveated event stream in various adverse environments such as low-light and motion scenes. Figure \ref{fig:rgb_in_low_light} shows the influence of motion and reduced ambient illumination on the ability of the RGB camera to take sharp snapshots of the text.
\\
The RGB camera struggles to produce sharp images due to increased exposure times in low-light scenes. With a cutoff at 30-50lux for which all OCR results return a 100\% WER and CER.
Our Event-Based OCR pipeline was able to achieve a WER of 8.3\% and CER of 2.5\% using the LLM OCR approach at only 30 Lux brightness, reconstructing a legible binary image of the read text.\\
%\begin{figure}[h]
%    \centering
%    \includegraphics[width=\linewidth]{images/Event Based OCR Result 30 Lux.png}
%    \caption{Event-based text reconstruction using our method at 30 Lux ambient brightness with only micromotions of head and hands while reading.}
%    \label{fig:event_based_reconstruction}
%\end{figure}\\

\begin{figure}[h!]
    \centering
    \begin{subfigure}[b]{0.45\textwidth}
    \centering
    \includegraphics[width=\linewidth]{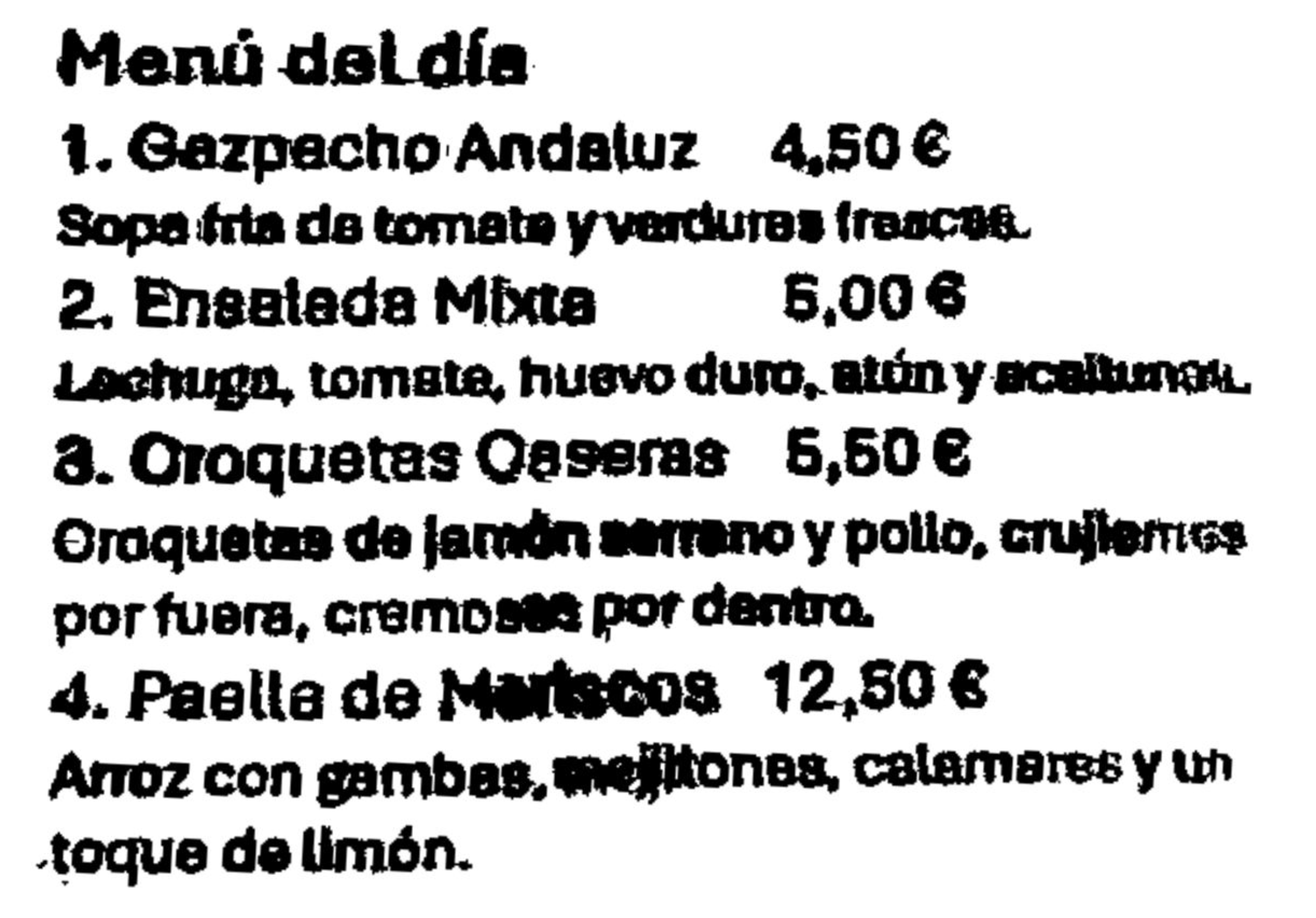}
    %\caption{Event-based text reconstruction using our method at 30 Lux ambient brightness with only micromotions of head and hands while reading.}
    \label{fig:event_based_reconstruction}
    \end{subfigure}
    \hfill
    \begin{subfigure}[b]{0.45\textwidth}
        \includegraphics[width=\textwidth]{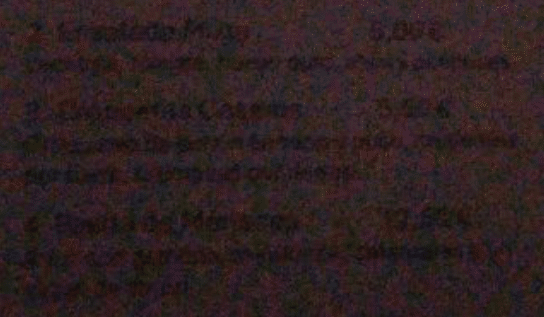}
        %\caption{Image 2}
        \label{fig:rgb_at_30_lux}
    \end{subfigure}
    \caption{Event-based text reconstruction on the top using our method and the RGB camera's view in the same recording on the bottom at 30 Lux ambient brightness with only micromotions of head and hands while reading.}
    \label{fig:side_by_side}
\end{figure}

While our event-based approach performs similarly in darkness as it does in daylight, the RGB camera's OCR performance varies strongly with illumination and head motion producing near-perfect OCR results in quasi-static, well-lit scenes.\\
Investigating the OCR performance WER and CER metrics for both traditional single snapshot RGB-based and our event-based foveated OCR approach shows the event-based approach outperforms the standard RGB-based approach both in dim-lit as well as high-motion scenes. RGB cameras such as the tested Meta Aria glasses integrated camera fail to recognize text below the region of 30-50 Lux due to motion blur incurred by a combination of micro motions, long exposure times, and image noise. Yet, the event-based approach is estimated to provide reasonable OCR output down to $\approx$~7 Lux (Twilight Brightness) thanks to the large dynamic range of event cameras. \\
Yet, due to the complex pipeline required to perform event-based OCR, the created binary representation of the real-world text does not match the quality of a high-resolution RGB image. This is due to small errors introduced during foveation, reconstruction, and image stitching which slightly increase the WER and CER of the event-based approach. This becomes obvious in well-lit and low-motion scenes, where the traditional RGB approach outperforms the event-based approach. Therefore, our event-based approach is well-suited to extend the operational envelope of egocentric OCR into high-motion and low-light scenes while not yet matching the performance of a traditional RGB approach in quasi-static, well-lit scenes.
%But due to the complex pipeline required to perform OCR on a foveated event stream, its representation of the read text never quite achieves the quality of a traditional RGB camera image which in turn hampers its WER and CER performance in well-lit and low-motion scenes compared to the traditional RGB approach. Therefore it is well suited to extend the operational envelope of traditional RGB cameras for egocentric OCR while nearly matching their performance in easier situations and could very well be used as a replacement.

% \cameraready{An important consideration when interpreting our results is the difference in physical and electrical characteristics between the RGB and event cameras. To account for the differing sensor sizes, we matched their angular resolution by using a 12 mm focal length lens on the event camera. Both cameras had a relative aperture (f-number) of F/2.4, ensuring that the total amount of light reaching each sensor was comparable. However, the light received per pixel differed due to their resolutions: the event camera has a resolution of 1280×720, while the RGB camera has a resolution of 2880×2880 across three channels. 
% % These differences in the sensor characteristics are particularly relevant for egocentric devices such as smart glasses, where physical constraints limit the size of embedded imaging components.
% The sensor size is particularly relevant for egocentric devices such as smart glasses, where physical constraints limit the dimension of embedded imaging components.
% }

\cameraready{An important consideration when interpreting our results is the difference in sensor size between the RGB and event cameras.
The sensor size is particularly relevant for egocentric devices such as smart glasses, where physical constraints limit the dimension of embedded imaging components.
The event camera has a pixel size of \SI{4.86}{\micro\metre} and a resolution of 1280×720, while the RGB camera has a pixel size of \SI{1.55}{\micro\metre} and a resolution of 2880×2880 across three channels.
To account for the differing sensor sizes, we matched their angular resolution by using a 12 mm focal length lens on the event camera. Both cameras had a relative aperture (f-number) of F/2.4, ensuring that the total amount of light reaching each sensor was comparable.
The fact that both sensors receive a comparable amount of light supports the validity of our results when comparing the two cameras under different ambient lighting conditions.}

% An aspect to consider in this analysis is the form factor of the cameras. 
% In our experiments, the pixel area of the event camera is approximately ten times larger than that of the RGB camera, this difference should be taken into account when interpreting the results presented in this section.
% This consideration is particularly relevant for egocentric devices, such as smart glasses, where space constraints limit the size of embedded sensors.

\subsection{Dedicated OCR vs. LLM OCR}
\label{sec:dedicated_ocr_vs_llm_ocr}
\begin{figure}[h]
    \centering
    \includegraphics[width=\linewidth]{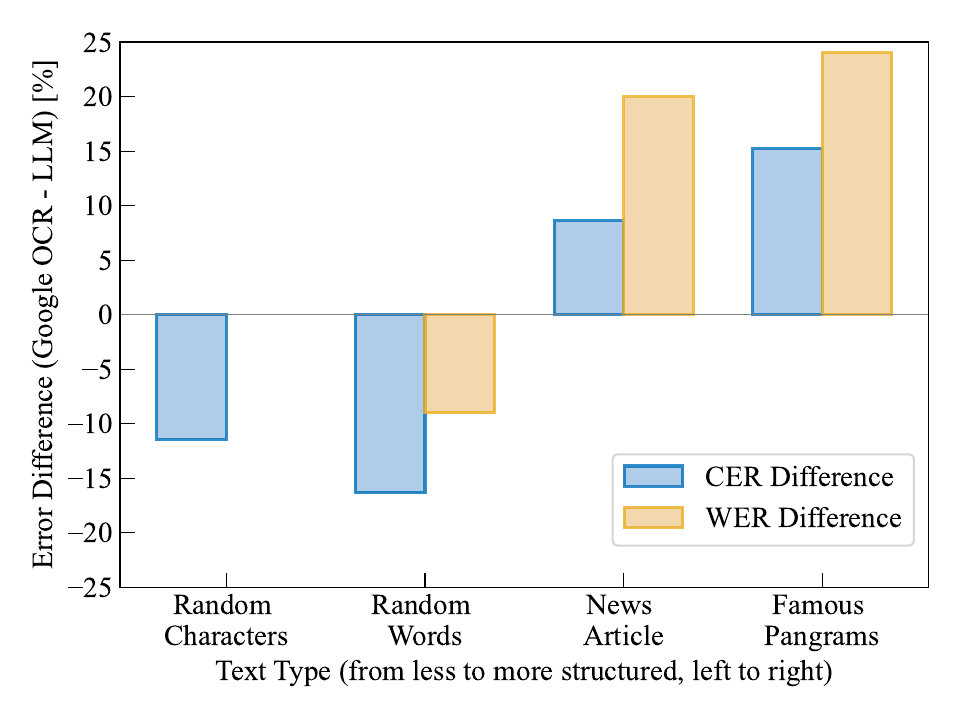}
    \caption{Four different levels of text coherencies were tested for both LLM-based and dedicated OCR methods. The Difference is being reported in WER (where possible) and CER.}
    \label{fig:LLM_vs_OCR}
\end{figure}

The dedicated Google Cloud OCR API and the LLM OCR approach based on OpenAI's ChatGPT-4o were compared using multiple text types of various coherency in non-adverse conditions. The goal was to evaluate the difference in OCR performance of both models based on the type of text input. This can be seen in Figure \ref{fig:LLM_vs_OCR}. The plotted difference in performance on the Y-axis for both WER and CER shows an increase in performance (and therefore a decrease in WER and CER) for LLM-based OCR for more structured text such as famous pangrams or a news article compared to the dedicated OCR software. The opposite is the case for less structured texts such as random words or random characters (which do not have a WER as there are no words). This shows the ability of an LLM-based OCR approach to outperform dedicated OCR software in transcribing text from images in situations with structured texts.\\
We compared dedicated cloud-based OCR applications to more recent multimodal LLMs with the ability to take images as direct input for OCR in various levels of text coherency investigating their WER and CER. Our findings show clear evidence of LLM-based OCR performing error correction based on their knowledge of sentence structures, and the discussed topics for coherent texts. Whereas for incoherent random word and random letter examples, the dedicated OCR solutions outperform LLM-based OCR. This is likely due to their specialized letter-recognizing capabilities. Overall, as an always-on smart glass AI agent will likely be LLM-based in the first place, LLM-based OCR could very well be the path to go in future wearable applications. This means it might not be required to send specific OCR queries to the cloud-based or edge-based multimodal LLM but rather include it in a continuous stream of, possibly foveated, information. 
\subsection{Bandwidth Reduction Results}
\label{sec:bandwidth_reduction_results}

\begin{figure}[h]
    \centering
    \includegraphics[width=\linewidth]{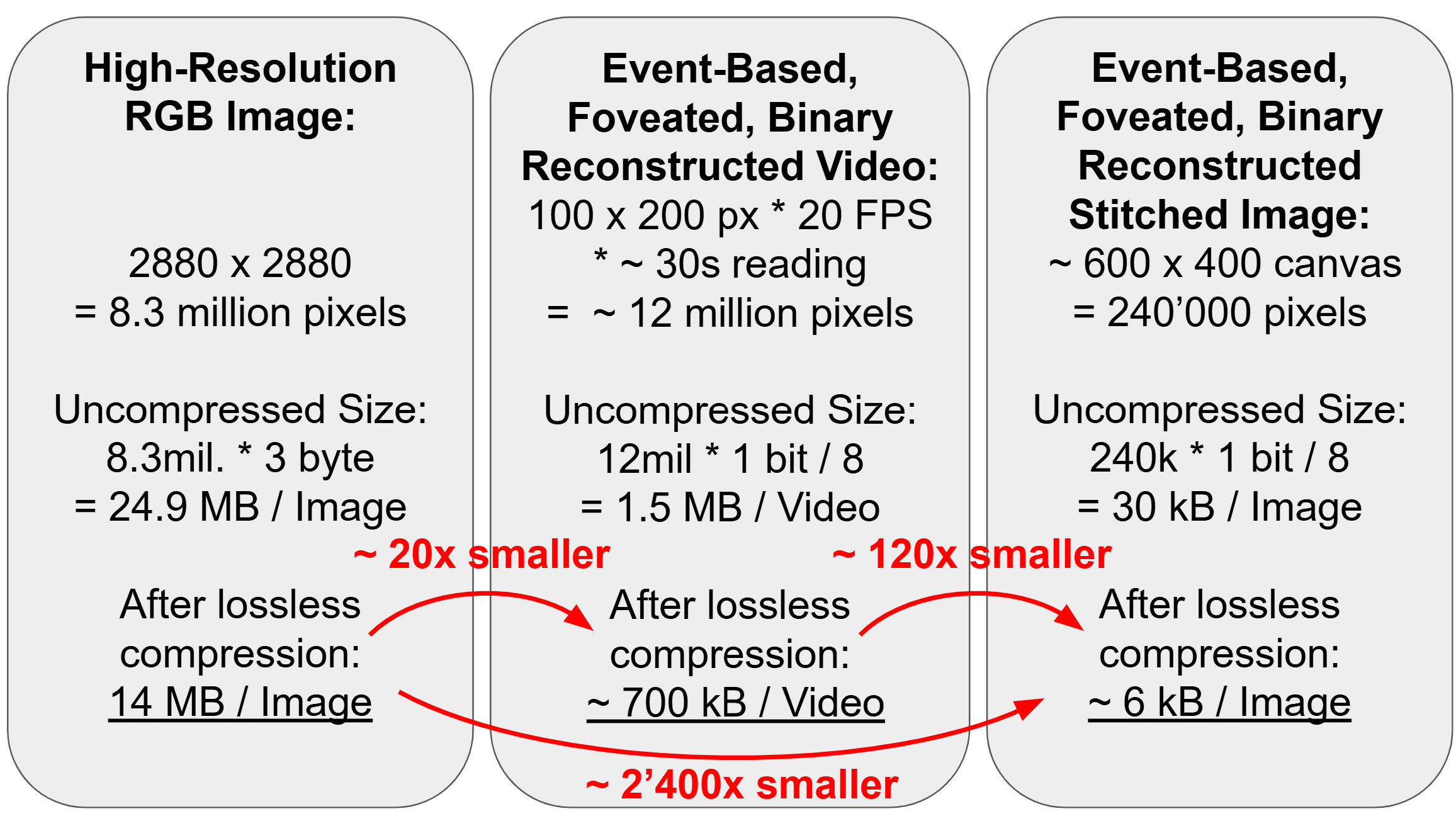}
    \caption{File size reduction of RGB-based OCR using a single image for transmission compared to different stages in our event-based OCR approach using event-stream foveation and binary image reconstruction.}
    \label{fig:file_size}
\end{figure}
%
%For JPG:  original PNG: 14MB, 100%JPG: 7.9Mb, 80% JPG: 1.29Mb, 60% JPG: 0.78Mb, 40% JPG: 0.5Mb, 20% JPG: 0.22Mb, 
Our approach shows a $\approx$ 20x reduction in filesize utilizing the foveated event stream with our binary reconstruction neural network to generate a binary foveated video stream instead of sending the single high-resolution RGB  image to the cloud for processing. Furthermore, there is $\approx$ 2'400x filesize reduction possible if hierarchical image stitching is performed on the smart glasses. This transforms the read text into a simple black-and-white image with reduced size before transferring it to the cloud for processing. This reduction can be seen in Figure \ref{fig:file_size}. 

The RGB image can be compressed into the JPG format to reduce its data size based on a specified quality level (indicated in parentheses): 7.9MB (100\%), 1.29MB (80\%), 0.78MB (60\%), 0.5MB (40\%), and 0.22MB (20\%). 
However, this compression process introduces data loss, which can negatively affect OCR performance.

\section{Discussion}
\label{sec:discussion}

\subsection{Bandwidth Reduction Implications}

As the target device is a low-power, wearable smart glass device, low-power and low computation is important to enable long battery life and prompt system response. Furthermore, the overarching goal of having an always-on AI assistant relies on a constant wireless connection to cloud-based services such as an LLM.
The main factors for battery-drain of the smart glasses are therefore extensive on-device computation and the transmission of large bandwidth items such as high-resolution images and videos~\cite{ragona2015}.
Therefore, next to a low WER and CER, a reduction in bandwidth is a key metric in reducing the power use and latency introduced during the transmission of the data to the cloud over 4G/5G or WiFi.
For this reason, the data bandwidth necessary to be transported from the wearable to the cloud is a key factor to reduce while keeping on-device computation as low as possible.

\begin{figure}[h]
    \centering
    \includegraphics[width=1\linewidth]{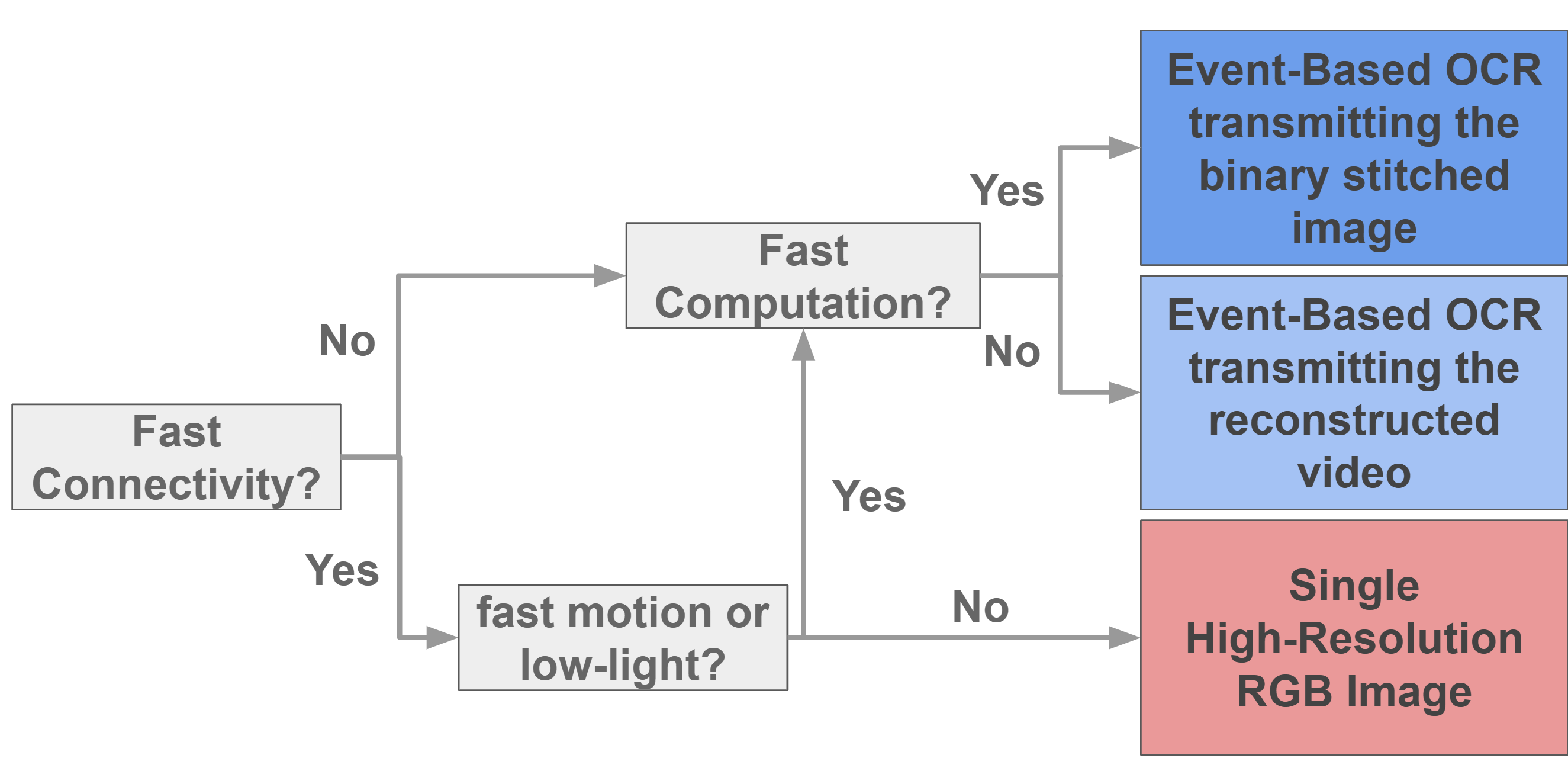}
    \caption{Flow diagram visualizing the strength of each investigated approach giving a guideline to the correct choice based on system parameters.}
    \label{fig:flow_diagram_approaches}
\end{figure}

Of course, this is highly dependent on the available computational power of the smart glasses and its power-consumption tradeoff versus transmitting more data.\\
As shown in Section \ref{sec:bandwidth_reduction_results}
our event-based, foveated, and binary reconstructed OCR approach shows great potential to reduce the file size and therefore the required bandwidth necessary to send the scene representation to a cloud-based server for OCR. Yet, this reduction in bandwidth comes with an increase in computation cost on the wearable itself, especially if image stitching is to be performed on the smart glasses. As shown in Figure \ref{fig:flow_diagram_approaches} two versions of our pipeline are investigated while the most likely solution for a power and computation-constrained system such as egocentric wearables (smart glasses) is to perform event-based recording, foveation, and binary reconstruction on the glasses. Thanks to the ability to stream foveated data in a continuous matter compared to the single snapshot RGB image approach relying on the transmission of a single large image file. This would allow us to further reduce the latency of a response to an OCR inquiry.
\section{Conclusion}
\label{sec:conclusion}

Based on the results found in this study, event-based cameras could be a viable alternative to the power-hungry and large bandwidth-requiring RGB-based cameras of current wearable egocentric vision systems. Especially, as in recent years event cameras have reached the angular resolution and size/pixel threshold to be an alternative to RGB cameras for egocentric OCR. We have seen plenty of benefits ranging from low-light and high-motion scene OCR performance to reduced latency, power consumption, and bandwidth thanks to the innovative approach of foveation and binary reconstruction of the egocentric event stream. This approach is likely also transferable to non-OCR-related tasks such as action recognition, segmentation, classification, or object detection enabling a personalized smart glass AI agent to run for longer while keeping the smart glasses small and light-weight enough for daily use. All while capturing more relevant and processing less redundant context from the environment compared to the use of high-resolution periodic RGB camera snapshots.
{
    \small
    \bibliographystyle{ieeenat_fullname}
    \bibliography{main}
}


\end{document}